\documentclass[11pt]{article}

\usepackage[a4paper,margin=1in]{geometry}
\usepackage{booktabs}
\usepackage{amsmath,amssymb}
\usepackage{graphicx}
\usepackage{hyperref}
\usepackage{url}

\title{An Agent-Oriented Pluggable Experience-RAG Skill for Experience-Driven Retrieval Strategy Orchestration}

\author{
zhangdutao\\
Macao Polytechnic University\\
\texttt{p2527418@mpu.edu.mo}
\and
Tian Liao\\
Macao Polytechnic University\\
\texttt{p2421086@mpu.edu.mo}
}

\date{}

\begin{document}

\maketitle

\begin{abstract}
Retrieval-augmented generation systems often assume that one fixed retrieval pipeline is sufficient across heterogeneous tasks, yet factoid question answering, multi-hop reasoning, and scientific verification exhibit different retrieval preferences. We present Experience-RAG Skill, an agent-oriented pluggable retrieval orchestration layer positioned between the agent and the retriever pool. The proposed skill analyzes the current scene, consults an experience memory, selects an appropriate retrieval strategy, and returns structured evidence to the agent. Under a fixed candidate pool, Experience-RAG Skill achieves an overall nDCG@10 of 0.8924 on BeIR/nq, BeIR/hotpotqa, and BeIR/scifact, outperforming fixed single-retriever baselines and remaining competitive with Adaptive-RAG-style routing. The results suggest that retrieval strategy selection can be productively encapsulated as a reusable agent skill rather than being hard-coded in the upper workflow.
\end{abstract}

\section{Introduction}

Retrieval-augmented generation (RAG) improves knowledge access in large language models by introducing external evidence before or during generation \cite{lewis2020rag,guu2020realm,izacard2021fid,borgeaud2022retro}. Existing work has explored query rewriting, hierarchical retrieval, active retrieval, corrective retrieval, self-reflective retrieval, and long-context retrieval \cite{ma2023rewrite,sarthi2024raptor,jiang2023active,yan2024crag,asai2023selfrag,jiang2024longrag}. However, many practical systems still assume that one fixed retrieval strategy is sufficient across all tasks.

This assumption is problematic in realistic agent settings. Factoid question answering, multi-hop retrieval, and scientific claim verification do not rely on the same retrieval behavior. A dense retriever may be strong for direct fact questions, while hybrid retrieval is often more stable for multi-hop and scientific evidence matching. For an agent, the challenge is therefore not only whether knowledge can be retrieved, but also how retrieval should be organized for the current task.

We address this problem by redefining Experience-RAG as an agent-oriented pluggable skill rather than a single retrieval method. This framing is related to recent work on tool-using and acting language agents \cite{yao2023react,schick2023toolformer,shen2023hugginggpt,patil2023gorilla,sumers2023cognitive}, but focuses specifically on retrieval strategy orchestration. The proposed Experience-RAG Skill acts as a middle layer between the agent and the retriever pool, with four core responsibilities: scene analysis, experience lookup, strategy routing, and result packaging. The current study focuses on a fixed candidate pool setting and investigates whether such a skill layer yields practical value even before open-world candidate onboarding is solved.

Our contributions are threefold. First, we formulate retrieval strategy selection as an agent skill orchestration problem. Second, we build an experience memory that records scene features, multi-retriever performance, and strategy margins. Third, we evaluate the proposed skill on heterogeneous retrieval tasks and compare it with both traditional and modern baselines.

\section{Experience-RAG Skill}

\subsection{Framework}

Given a query $q$, dialogue history $h$, task metadata $m$, and a candidate retriever pool $\mathcal{R}$, Experience-RAG Skill first constructs a structured scene representation $s$, then routes the query using experience memory $M$, and finally returns a standardized retrieval package:
\begin{equation}
s = Analyze(q, h, m), \quad r^* = Route(s, M), \quad o = RetrieveAndPackage(q, r^*).
\end{equation}

The framework contains six modules: Skill Interface, Scene Analyzer, Experience Memory, Strategy Router, Retriever Pool, and Result Packager. The agent only interacts with the unified skill interface, while the lower retrieval methods remain encapsulated inside the pool.

\begin{figure}[t]
\centering
\includegraphics[width=\linewidth]{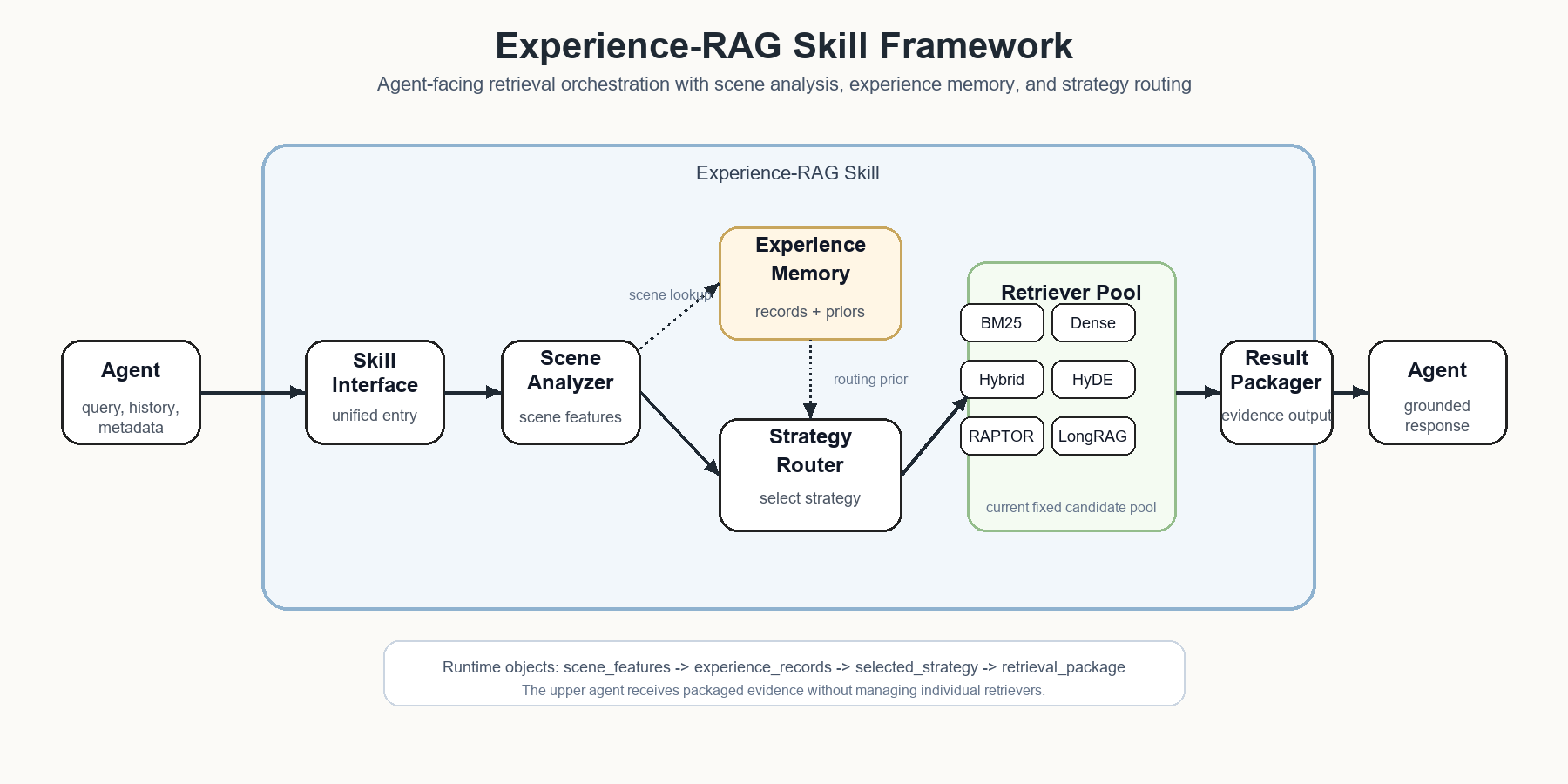}
\caption{Overall framework of Experience-RAG Skill. The skill is positioned between the agent and the retriever pool. It transforms the incoming query, history, and metadata into structured scene features, consults reusable experience records, selects a retrieval strategy, invokes the candidate retriever pool, and returns a packaged evidence object to the agent.}
\label{fig:framework}
\end{figure}

\subsection{Experience Memory and Routing}

The experience memory stores complete records rather than only the best-strategy label:
\begin{equation}
e = (scene\_features, score\_vector, best\_strategy, best\_margin).
\end{equation}

In the current version, scene features include task type, domain, context length, question complexity, query style, and document structure. These features support both rule-based and learned routing.

Our strongest current setting is rule-based routing. It maps \texttt{direct} tasks to \texttt{dense}, and both \texttt{multi\_hop} and \texttt{scientific} tasks to \texttt{hybrid\_rrf}. Learned routing variants were also explored, but under the present experience scale they did not outperform the rule-based baseline.

\subsection{Difference from Adaptive-RAG}

Adaptive-RAG-style methods focus on complexity-driven strategy selection. In contrast, Experience-RAG Skill focuses on an agent-facing orchestration layer with explicit experience reuse, unified interfaces, and structured evidence packaging. This difference is important because our goal is not only to choose a retriever, but also to make retrieval strategy selection reusable as an agent capability.

\section{Evaluation}

\subsection{Setup}

We evaluate on three public retrieval benchmarks: BeIR/nq, BeIR/hotpotqa, and BeIR/scifact, using 120 queries from each dataset and sampled candidate corpora. We compare against \texttt{bm25}, \texttt{rewrite\_bm25}, \texttt{dense}, \texttt{hybrid\_rrf}, and \texttt{experience\_rag}. We report Recall@10, MRR@10, and nDCG@10.

\subsection{Main Results}

Table~\ref{tab:overall} shows that Experience-RAG Skill achieves the best overall retrieval quality, with nDCG@10 of 0.8924, Recall@10 of 0.9428, and MRR@10 of 0.9006. It outperforms the strongest fixed single-method baseline \texttt{hybrid\_rrf} (0.8802 nDCG@10) as well as \texttt{dense} (0.8627), \texttt{bm25} (0.8426), and \texttt{rewrite\_bm25} (0.8412). At the dataset level, Experience-RAG ties with the strongest fixed strategy on each dataset while outperforming all fixed strategies on the mixed workload, indicating that its main advantage lies in heterogeneous orchestration rather than dominance on every single subtask.

\begin{table}[t]
\centering
\caption{Overall retrieval performance.}
\label{tab:overall}
\begin{tabular}{lccc}
\toprule
Method & Recall@10 & MRR@10 & nDCG@10 \\
\midrule
BM25 & 0.9012 & 0.8556 & 0.8426 \\
Rewrite-BM25 & 0.8956 & 0.8566 & 0.8412 \\
Dense & 0.9136 & 0.8819 & 0.8627 \\
Hybrid RRF & 0.9428 & 0.8846 & 0.8802 \\
Experience-RAG Skill & \textbf{0.9428} & \textbf{0.9006} & \textbf{0.8924} \\
\bottomrule
\end{tabular}
\end{table}

\subsection{Modern Baseline Comparison}

We further expand the candidate pool with \texttt{HyDE}, \texttt{RAPTOR-style}, \texttt{LongRAG-style}, and \texttt{Adaptive-RAG-style}, covering query expansion, hierarchical retrieval, long-context retrieval, and adaptive routing \cite{gao2022hyde,sarthi2024raptor,jiang2024longrag,jeong2024adaptive}. Table~\ref{tab:modern} shows that Adaptive-RAG-style is slightly better overall (0.8934 vs.\ 0.8924 nDCG@10), but the gap is very small. In contrast, HyDE, LongRAG-style, and RAPTOR-style are clearly weaker under the current sampled-corpus and retrieval-only setting. These results suggest that Experience-RAG Skill is already competitive with the strongest modern routing baseline while providing a clearer agent-facing systems boundary.

\begin{table}[t]
\centering
\caption{Comparison with modern baselines.}
\label{tab:modern}
\begin{tabular}{lccc}
\toprule
Method & Recall@10 & MRR@10 & nDCG@10 \\
\midrule
Adaptive-RAG-style & \textbf{0.9428} & \textbf{0.9019} & \textbf{0.8934} \\
Experience-RAG Skill & \textbf{0.9428} & 0.9006 & 0.8924 \\
HyDE & 0.8983 & 0.8533 & 0.8326 \\
LongRAG-style & 0.7399 & 0.7407 & 0.7095 \\
RAPTOR-style & 0.7155 & 0.7223 & 0.6841 \\
\bottomrule
\end{tabular}
\end{table}

\section{Brief Discussion}

The current results support three conclusions. First, retrieval strategy selection can be productively encapsulated as an agent skill rather than being embedded in the upper workflow. Second, experience-driven orchestration is effective on heterogeneous retrieval workloads. Third, the skill remains competitive after modern baselines are added to the same fixed candidate pool.

At the same time, the current version has clear limitations. The experiments are based on sampled rather than full-scale corpora, learned routing does not yet outperform the rule-based baseline, and the current implementation still assumes a fixed candidate pool. Moreover, the present workflow case analysis is qualitative and does not yet constitute a full end-to-end interactive agent benchmark. These limitations make the current version more appropriate as a short paper or preprint than as a full systems paper.

\section{Conclusion}

We introduced Experience-RAG Skill, an agent-oriented pluggable retrieval orchestration layer that combines scene analysis, experience memory, and strategy routing. Under a fixed candidate pool, the proposed skill outperforms fixed single-retriever baselines and remains competitive with Adaptive-RAG-style routing on heterogeneous retrieval tasks. The study suggests that retrieval strategy selection is a reusable agent capability in its own right. Future work will extend this line toward dynamic candidate onboarding, capability-level routing, and full end-to-end agent evaluation.

\appendix
\section{Supplementary Experiments}

\subsection{Ablation Study}

To better understand where the gains of Experience-RAG Skill come from, we also evaluated several ablated variants. The full Experience-RAG configuration achieves nDCG@10 of 0.8924, whereas fixing the strategy to \texttt{hybrid\_rrf} yields 0.8802, fixing it to \texttt{dense} yields 0.8627, and fixing it to \texttt{bm25} yields 0.8426. These results indicate that the main advantage comes from task-aware orchestration rather than from a single strong retriever.

\begin{table}[h]
\centering
\caption{Ablation results.}
\label{tab:ablation-short}
\begin{tabular}{lccc}
\toprule
Variant & Recall@10 & MRR@10 & nDCG@10 \\
\midrule
Full Experience-RAG & \textbf{0.9428} & \textbf{0.9006} & \textbf{0.8924} \\
Fixed Hybrid RRF & 0.9428 & 0.8846 & 0.8802 \\
Fixed Dense & 0.9136 & 0.8819 & 0.8627 \\
Fixed BM25 & 0.9012 & 0.8556 & 0.8426 \\
\bottomrule
\end{tabular}
\end{table}

\subsection{Learned Routing Attempts}

We additionally tested learned routing variants in order to examine whether the experience memory can support automatic router learning. A hard-classification router reached nDCG@10 of 0.8778, while a score-regression router reached 0.8627. Both remained below the rule-based variant. Under the current scale of the experience memory, these results suggest that learned routing is feasible but not yet mature enough to replace the stronger rule-based skill.

\subsection{Preliminary Agent Workflow Observation}

We also carried out a qualitative workflow inspection with three representative tasks: factoid question answering, multi-hop retrieval, and scientific claim verification. Across these examples, Experience-RAG Skill consistently made the routing decision explicit and inspectable, whereas fixed-retriever agents lacked a dedicated mechanism for selecting retrieval behavior. This observation is supportive but still qualitative, and should not be interpreted as a full end-to-end agent benchmark.

\end{document}